\documentclass{article}
\usepackage{ijcai22}
\usepackage{natbib}

\usepackage{times}

\usepackage{soul}

\usepackage[hidelinks]{hyperref}
\usepackage[utf8]{inputenc}
\usepackage[small]{caption}
\usepackage{url}
\usepackage{graphicx}
\usepackage{amsmath}
\usepackage{booktabs}
\usepackage{latexsym}
\usepackage{subcaption} 
\usepackage{color,soul}
\usepackage[normalem]{ulem}
\usepackage{microtype}
\usepackage{algorithm}
\usepackage{algorithmic}
\usepackage{comment}
\usepackage{todonotes}
\usepackage{color,soul}
\usepackage{graphicx}
\usepackage{pbox}
\usepackage{subcaption}
\usepackage{epstopdf}
\usepackage{xstring}
\usepackage{algorithm}
\usepackage{algorithmic}
\usepackage[normalem]{ulem}
\usepackage{booktabs}
\useunder{\uline}{\ul}{}
\usepackage{comment}
\usepackage{wrapfig}
\usepackage{blindtext}
\usepackage{booktabs}
\usepackage{multirow}

\newcommand\tab[1][0.3cm]{\hspace*{#1}}

\title{Detecting and Understanding Harmful Memes: A Survey}

\author{
Shivam Sharma$^{2,7}$\and
Firoj Alam$^1$\and
Md. Shad Akhtar$^{2}$\and
Dimitar Dimitrov$^4$\and\\
Giovanni Da San Martino$^5$\and
Hamed Firooz$^6$\and
Alon Halevy$^6$\and
Fabrizio Silvestri$^3$\and\\
Preslav Nakov$^1$\And
Tanmoy Chakraborty$^{2}$
\affiliations
$^1$Qatar Computing Research Institute, HBKU, Qatar\\ 
$^2$IIIT-Delhi, India\\ 
$^3$Sapienza University of Rome, Italy\\ 
$^4$Sofia University, Bulgaria\\ 
$^5$University of Padova, Italy\\ 
$^6$Facebook AI, USA\\ 
$^7$Wipro AI Labs, India\\
\emails
\{fialam,pnakov\}@hbku.edu.qa, \{shivams,tanmoy,shad.akhtar\}@iiitd.ac.in, fsilvestri@diag.uniroma1.it, 	
mitko.bg.ss@gmail.com, dasan@math.unipd.it, \{mhfirooz,ayh\}@fb.com
}

\begin{document}

\maketitle

\begin{abstract}
The automatic identification of harmful content online is of major concern for social media platforms, policymakers, and society. Researchers have studied textual, visual, and audio content, but typically in isolation. Yet, harmful content often combines multiple modalities, as in the case of memes, which are of particular interest due to their viral nature. With this in mind, here we offer a comprehensive survey with a focus on \textit{harmful memes}. Based on a systematic analysis of recent literature, we first propose a new typology of harmful memes, and then we highlight and summarize the relevant state of the art. One interesting finding is that many types of harmful memes are not really studied, e.g.,~such featuring self-harm and extremism, partly due to the lack of suitable datasets. We further find that existing datasets mostly capture multi-class scenarios, which are not inclusive of the affective spectrum that memes can represent. Another observation is that memes can propagate globally through repackaging in different languages and that they can also be multilingual, blending different cultures. We conclude by highlighting several challenges related to multimodal semiotics, technological constraints, and non-trivial social engagement, and we present several open-ended aspects such as delineating online harm and empirically examining related frameworks and assistive interventions, which we believe will motivate and drive future research.
\end{abstract}

\section{Introduction}
\label{sec:introduction}

Social media have enabled individuals to freely share content online. While this was a hugely positive development as it enabled free speech, it was also accompanied by the spread of harm and hostility \citep{brooke-2019-condescending,joksimovic-etal-2019-automated}.

Hate speech \citep{fortuna2018survey}, 
offensive language \citep{Offensive:Type:Target:2019,zampieri2020semeval},
abusive language \citep{mubarak2017abusive}, propaganda \citep{EMNLP19DaSanMartino}, cyberbullying \citep{van-hee-etal-2015-detection}, cyber-aggression \citep{kumar2018benchmarking}, and other kinds of harmful content \citep{pramanick-etal-2021-momenta-multimodal}\footnote{{\scriptsize \textcolor{red}{\textbf{Disclaimer:}}
\textcolor{red}{
This paper contains content that may be disturbing to some readers.}}} have become prominent online.
Such content can target users, communities (e.g., minority groups), individuals, and companies. Social media have defined various categories of harmful content that they do not allow on their platforms \citep{DBLP:journals/corr/abs-2009-10311,nakov2021detecting}, and various categorizations of such content have also come from the research community \citep{banko-etal-2020-unified,pramanick-acl}.

Social media content is often multimodal, combining text, images, and/or videos. In recent years, \textit{Internet memes} have emerged as a prevalent type of content shared on social media. A meme is ``a group of digital items sharing common characteristics of content, form, or stance, which were created by associating them and were circulated, imitated, or transformed via the Internet by many users''~\citep{shifman2013memes}. Memes typically consist of images containing some text \citep{shifman2013memes,suryawanshi-etal-2020-dataset,suryawanshi-etal-2020-multimodal}. The design used in memes is typically humorous, but they are often harmful.

There has been a lot of work on detecting content that is harmful or otherwise violates the terms of service of online platforms \citep{alam2021survey,nakov2021detecting,pramanick-acl,pramanick-etal-2021-momenta-multimodal}. This includes detecting hateful users on Twitter \citep{ribeiro2018characterizing}, understanding the virality patterns of memes \citep{chenimageVIRAL2021}, detecting offensive and non-compliant content/logos in product images \citep{Gandhi2020ScalableDO}, spotting hate speech in videos and other modalities \citep{Gomez2020exploring,ChingHatespeechVid2020}, as well as detecting fine-grained propaganda techniques in memes~\citep{dimitrov2021detecting}, among others. More generally, some of the latest surveys on specific aspects of violating content have been on detecting fake news~\citep{thorne-vlachos:2018:C18-1,Islam2020,Kotonya2020}, disinformation \citep{alam2021survey,Survey:2022:Stance:Disinformation}, misinformation \citep{Survey:2021:AI:Fact-Checkers,Survey:2021:Media:Factuality:Bias}, rumours \citep{bondielli2019survey}, propaganda \citep{da2020survey}, memes \citep{afridi2021multimodal}, hate speech \citep{fortuna2018survey,Schmidt2017survey}, cyberbullying \citep{7920246}, and offensive content \citep{husain2021survey}. 

Our survey focuses on detecting and analyzing harmful memes, i.e.,~\textit{multimodal units consisting of an image and embedded text that have the potential to cause harm to an individual, an organization, a community, or society in general}.

\begin{figure}[t!]
\includegraphics[width=\columnwidth]{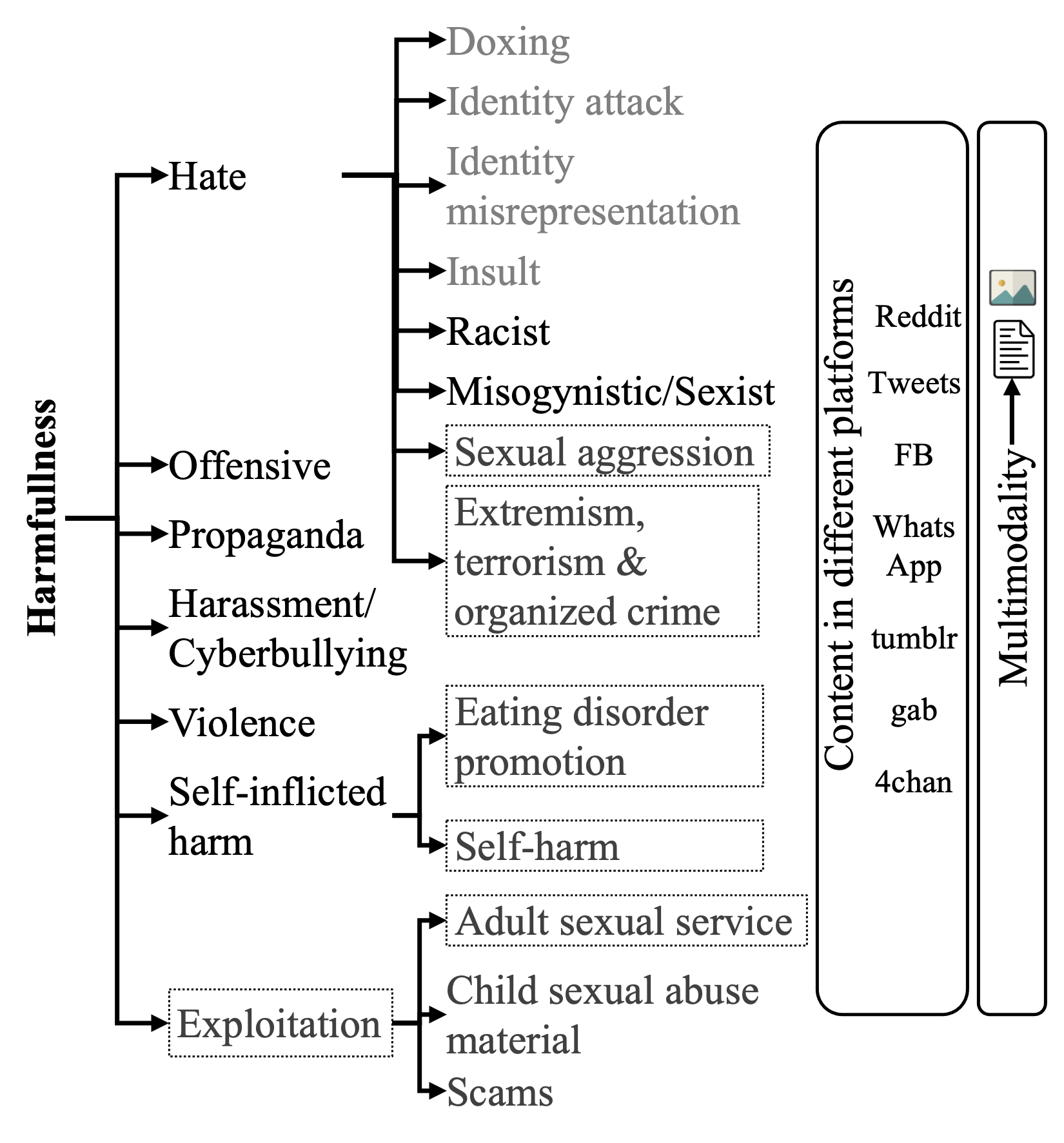}
\centering
\caption{Typology of harmful memes. We show in \emph{gray color} the categories for which we found no memes and no research publications; dotted boxes indicate that this type of memes exist, but we found no publications trying to detect it.}
\label{fig:harmful_content_tax}
\end{figure}

\begin{figure}[t!]
\centering
\resizebox{\columnwidth}{!}{
\includegraphics[width=0.5\textwidth]{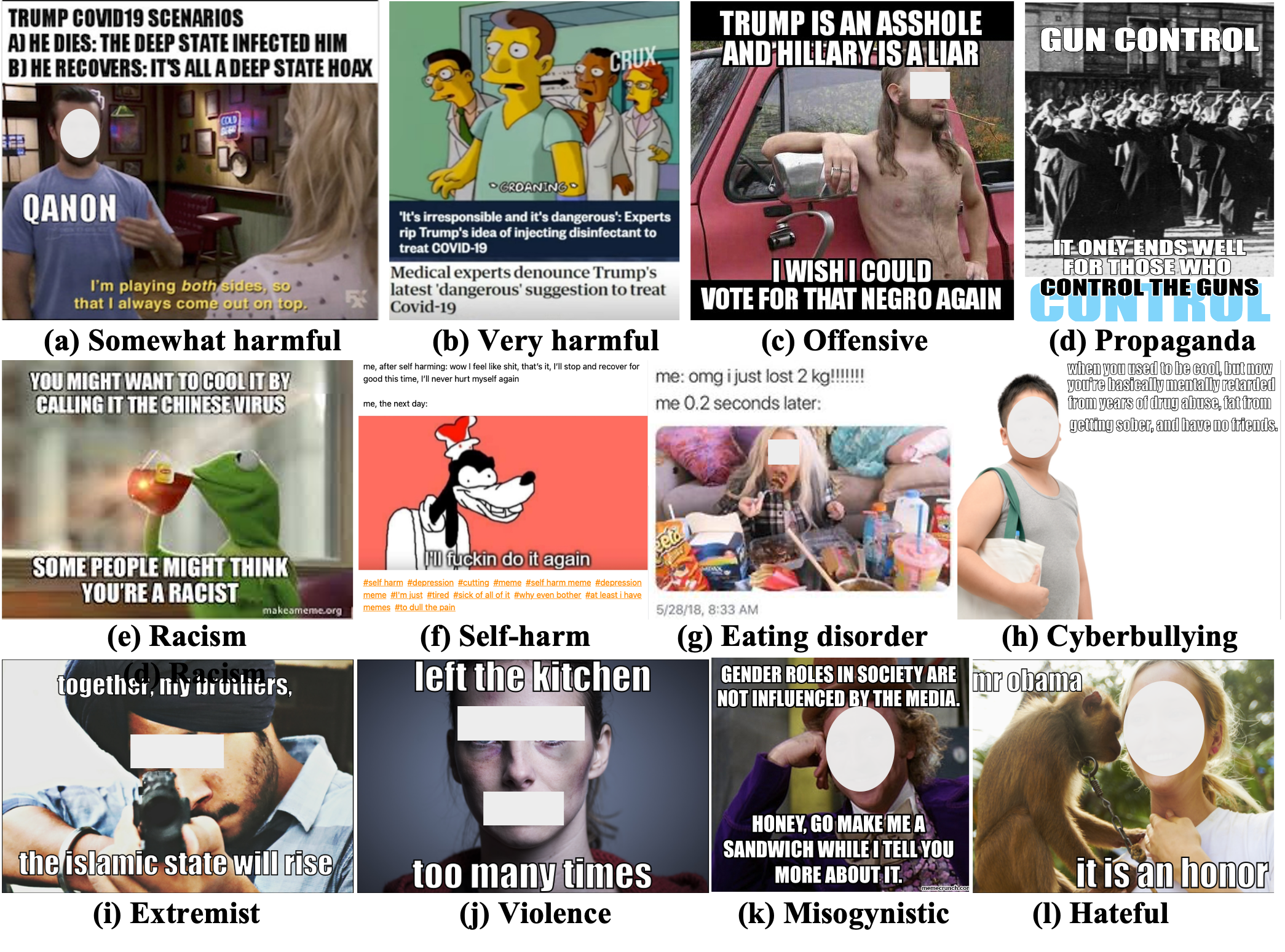}}
\caption{
Examples of different types of harmful memes.  
}
\label{fig:harmful_memes_examples}
\end{figure}

Figure~\ref{fig:harmful_content_tax} shows our typology of harmful memes, which we defined based on an extensive literature survey; examples of different types of harmful memes are shown in Figure~\ref{fig:harmful_memes_examples}. Below, we discuss various aspects of the typology, as well as multimodality, multilinguality, cultural influences, and global propagation through repackaging. We further highlight key issues including the need for fine-grained analysis, the complex abstraction of the memes, and the challenges of the subjectivity of the annotations and of multimodal learning.

\section{Harmful Memes}
Below, we present our new typology for harmful content on social media, with special focus on meme dissemination. We believe that it would help contextualize the scope not only for ongoing investigations, but also for future research. Figure~\ref{fig:harmful_content_tax} depicts this typology, which is inspired but differs from what was proposed in previous work \citep{banko-etal-2020-unified,nakov2021detecting,pramanick-acl}.

For example, \citet{banko-etal-2020-unified} categorized misinformation as ideological harm, which we excluded from our typology as misinformation is not always harmful. Similarly, while the intent of disinformation is harmful by definition, we do not specifically include it in our typology as most of our sub-categories (e.g.,~\emph{hate} and \emph{violence}) fall under disinformation~\citep{alam2021survey}. Similarly, some of our sub-categories (e.g.,~\emph{doxing} and \emph{identity attack}) fall under malinformation. Figure~\ref{fig:harmful_content_tax} highlights the categories with grey-coloured text in a dotted box for which we could not find any studies, even though they are prominent in social media: for example, a query in a major search engine using the keywords from Figure~\ref{fig:harmful_content_tax} will return many memes expressing the respective type of harm~\citep{sabat2019hate}.

\subsection{Types of Harmful Memes}

\subsubsection{\textbf{I: Hate}}
Studies on hate speech detection have focused primarily on textual content \citep{fortuna2018survey}, and less on the visual modality \citep{ChingHatespeechVid2020}, with limited research focus on memes \citep{kiela2020hateful,9455994}. An enabling effort in this respect was the \emph{Hateful Memes Challenge} \citep{kiela2020hateful}, which aimed to identify the targeted protected categories (e.g.,~\emph{race} and \emph{sex}) and the type of attack (e.g.,~\emph{contempt} and \emph{slur}) in memes~\citep{zia-etal-2021-racist}. The best system in the competition used different unimodal and multimodal pre-trained models such as VisualBERT \citep{li2019visualbert} VL-BERT \citep{su2019vl}, UNITER \citep{chen2019uniter}, VILLA \citep{gan2020large}, and ensembles thereof \citep{kiela2021hateful}. Using the same dataset, \citet{9455994} proposed a novel method by incorporating image captioning and data augmentation. The shared task on hateful memes at WOAH 2021 introducied new labels and tasks, which \citet{zia-etal-2021-racist} addressed using state-of-the-art pre-trained visual and textual representations along with logistic regression. There have also been efforts to detect the specific \emph{protected categories} being targeted. Below, we elaborate on two such major protected categories: racist and misogynistic/sexist, which are most common in hateful memes in social media. 

\tab {\textbf{I.A: Racist:}}
The race is one such protected category that has multi-dimensional aspects in which a systematic out-casting takes place within social, economic, and cultural ecosystems. It is defined as,\footnote{\url{https://dictionary.cambridge.org/dictionary/english/racism}} ``\textit{Policies, behaviours, rules, etc. that result in a continued unfair advantage to some people and unfair or harmful treatment of others based on race}.'' Memetic racism mostly leverages the following:

\tab \tab {\textbf{I.A.a: Physical Appearance:}}
Online racism was found to be prominently based on physical appearances through memes. Research studies used keyword-based scraping of memes from platforms such as Gab, Twitter, 4chan, etc., followed by an in-depth qualitative discussion of the characteristics of online discourse, and supported by thematic analysis. \citet{WILLIAMS2016424} investigated the correlation between the offline racial experiences and online perception of racism, where user feedback from white people and people of colour was obtained for understanding the differences in the perception of racism. Their findings suggested a higher likelihood of perceiving racism online, primarily by offline victims.

One of the classic scenarios of demeaning people of color and camouflaging systematic racism, also referred to as \emph{color-blindness racism} \citep{Yoon2016WhyII}, against African-Americans is the usage of standard meme templates that primarily target black NBA athletes whilst juxtapositioning against white men from the NFL, thereby promoting white supremacy \citep{Nikolas2016}. This is also exemplified within racism by non-indigenous Australians against Aboriginals, which primarily leverages skin tone, stereotypes, and phenotypical characteristics. These memes use either slur/racist words like \emph{abo} and \emph{abbos} or crop facial depictions of aboriginals to convey white supremacy and vilification \citep{NatourAboriginals2020}.  


\tab\tab{\textbf{I.A.b: Ethnicity:}}
Ethnocultural aspects are prominent online, as users from various cultural backgrounds share a common platform for exchanging ideas. \citet{Fairchild2020ItsFB} presented a generic thematic analysis of nine codesets focusing on \emph{race and ethnicity, slurs and language, stereotypes, typology, politics, and culture}, followed by a contextual analysis of the racist discourse and associated tags. \citet{tuters_para2020} presented a qualitative perspective of the prevalence of \emph{triple parenthesis} meme promoting hostility against Jews on 4chan's /pol/. \citet{zannetou2018}  empirically analyzed (\emph{i})~the spread of anti-Semitic memes like the \emph{Happy Merchant} meme via semantic embeddings, and (\emph{ii})~the temporal influence that fringe online users have towards their normalization into mainstream media using the Hawkes process and change-point analysis. They highlighted the use of derogatory slang words, nationalism, conspiracy theories grounded in biblical literature, and hatred towards Jews, encoded via visual-linguistic instruments \citep{Fairchild2020ItsFB}. \textit{Floating signifiers} (e.g.,~the \emph{Pepe the Frog} meme) along with the adversarial language games \citep{tuters_para2020}, lend themselves as versatile and highly accessible platforms for malevolence. As mentioned earlier, social media platforms are instrumental in propagating various types of harmful memes. \citet{zannetou2018} studied 4chan's /pol/ as the major unidirectional spreader of the \emph{Happy Merchant} meme, among many other platforms. 

\citet{Fairchild2020ItsFB} highlighted the pivotal role that gamer communities play in facilitating the spread of highly racist content against the generic ones that enable moderately racist content. From computation studies' viewpoint, \citet{Chandra_Jew2021} emphasized optimal encoding of different modalities using models like ResNET152~\citep{he2016deep} and RoBERTa~\citep{liu2019roberta}, along with Multimodal Fusion Architecture Search (MFAS), yielding 0.71 and 0.90 F1 score for Twitter and Gab datasets, respectively, suggesting greater propensity for multimodality in the latter.   

\tab {\bf I.B: Misogynistic/Sexist:}
Misogyny and sexism against women have grown a foothold within social media communities, reinvigorating age-old patriarchal establishments of baseless name-calling, objectifying their appearances, and stereotyping gender roles, which has been explored in the literature~\citep{france_mys2021}. This is especially fueled by the cryptic use of humor disguised as sexism via memes. Qualitative analysis involving the identification of the dominant themes present within sexist memes, followed by their detailed interpretation was done via adjectival assessment and with a focus on themes like \emph{technological privilege}, \emph{others}, \emph{dominance of patriarchy}, \emph{gender stereotypes}, and \emph{women as manipulators} in \citep{Drakett2018,Siddiqi2018}. Further analysis by the same authors showed use of derogatory language in these memes, accompanied by the depiction of confident, strong and poised women, essentially suggesting the threat perceived by sexist and chauvinistic people. When considered for a more extensive set of online memes, such imagery could also be present in non-sexist memes, which highlights the importance of the textual modality. This is further corroborated for sexist meme detection in \citep{fersini2019detecting}, where textual cues with a late-fusion strategy yields an F1 score of 0.76, highlighting the efficacy of distinctly modeling textual cues for such scenarios.

\subsubsection{II: Offensive Memes}

Offensive content intends to upset or embarrass people by being rude \citep{suryawanshi-etal-2020-multimodal}. Several studies have focused on content and implicit offensive analogies within memes. Some leveraged unimodal \citep{9402406} and multimodal information \citep{suryawanshi-etal-2020-multimodal}, yielding and F1 score of 0.71 and accuracy of 0.50 towards investigating simple encoder and early fusion-based strategies for classifying offensive memes, whilst employing techniques like stacked LSTM/BiLSTM/CNN (Text) along with VGG-16~\citep{simonyan2014very} towards multi-modality inclusion. Addressing contextualisation, \citet{AOMD2021} used analogy-aware multimodality (using ResNet50~\citep{he2016deep}, GloVe-based LSTM) and attentive multimodal analogy alignment via supervised learning, while incorporating the contextual discourses, yielding 0.72 and 0.69 accuracy for Reddit- and Gab-based datasets, respectively. \citet{9582340} extended this study via a graph neural network approach towards multimodal entity extraction (KMEE) by leveraging common-sense knowledge towards detecting offensive memes, which led to 1\% accuracy enhancement for both scenarios. This suggests the importance of contextual and commonsense knowledge for modeling the offensive content in memes.

\subsubsection{III: Propaganda}
\label{sssec:propaganda}
Harmful propaganda memes are prominent in online fora that promote \emph{xenophobia, racial strangers, anti-semitic, self-promotion, and anti-feminist/LGBTQ} \citep{nordinNeoNaziAltright2021,Dafaure2020TheM}. The memetic language involves similar styles, symbolism and iconography for contrasting inclinations \citep{studamerhumor2019} towards recruitment and promoting violent racial supremacy \citep{proudboys2018,askaniusMurder2021}. 

\citet{mittos2019and} investigated \emph{genetic testing} discourse, involved in establishing racial superiority and far-right ideologies, by studying resulting correlations using topic modeling, contextual semantics, toxic content analysis, and pHash to characterize the visual cues in memes. Recently, a novel multimodal multi-label fine-grained propaganda detection task from memes was proposed \citep{dimitrov2021detecting}, including a shared task at SemEval-2021 \citep{SemEval2021-6-Dimitrov}, with a focus on fine-grained propaganda techniques in text and the entire meme, confirming the importance of multimodal cues.

\subsubsection{IV: Harassment/Cyberbullying}
\label{sssec:harassment_cyberbullying}
The terms \emph{harassment} and \emph{cyberbullying},  are often used interchangeably in the literature. The difference between them is subtle: when the bullying behaviour is directed at the target based on race, skin color, religion, sex, age, disability, and nationality, it is also defined as harassment. In the past decade, there have been significant research efforts and initiatives by policymakers and social media platforms to address the issue of online harassment and cyberbullying, as it has been leading to suicides and psychological distress \citep{rosa2019automatic}. The study of PAW research highlights the increase of harassment over time, most of which happens through social media platforms \citep{vogels2021state}. The automatic detection of such harassment or cyberbullying content has been a significant focus for computational social science. \citet{rosa2019automatic} systematically reviewed for automatic cyberbullying detection and listed the available datasets, methodologies, and state-of-art performance. They also provided an operational definition exemplifying cyberbullying while delineating annotation guidelines and agreement measures, along with ethical aspects. Besides focusing on the textual modality, \citet{Hosseinmardi2015cyberbull} also investigated Instagram images and their associated comments for detecting cyberbullying and online harassment. They manually curated a dataset of 998 samples, including images and their associated comments. Interestingly, they noted that 48\% of the posts with loaded language were not labelled as cyberbullying. \citet{singh2017toward} also investigated cyberbullying detection using the same dataset and observed that the image and the text modalities complement each other. Despite the continued use of multimodal content and memes for cyberbullying, we could not find any significant efforts towards its automated detection. However, \emph{name-calling}, which is a prominent tool for cyberbullying, has been explored for propaganda detection \citep{SemEval2021-6-Dimitrov}.

\subsubsection{V: Violence}
\label{sssec:violence}
Violence is defined as \emph{``the intentional use of physical force or power, threatened or actual, against oneself, another person, or against a group or community, that either result in or have a high likelihood of resulting in injury, death, psychological harm, maldevelopment or deprivation''}~\citep{krug2002world}. 

There has been a lot of research in the past decades focusing on multimodal violence detection in surveillance videos \citep{ramzan2019review,yao2021survey}, based on video and audio modalities \citep{10.1145/2502081.2502187}. Another line of research investigated the \emph{threat of violence}~\citep{banko-etal-2020-unified} in comments on YouTube videos and Wikipedia \citep{wulczyn2017ex}. In the existing literature, the automatic detection of violent memes has been studied in various contexts, e.g.,~detecing hateful memes \citep{kiela2020hateful}. 
Yet, we could not find any work specifically focusing on violent meme detection.

\subsubsection{VI: Self-Inflicted Harm}
\label{sssec:self_inflicted_harm}
Self-inflicted harm includes different forms of harmful behaviour, such as self-injury, eating disorders, suicide and other self-harming behaviors \citep{seko2018self,banko-etal-2020-unified,SIGIR:2022:suicide}. It can be both physical and psychological, and most people self-injure to cope with negative emotions, punish themselves or solicit help from others \citep{seko2018self}. Studies suggest that social media (e.g.,~Tumblr) have been the hotbed for featuring such self-harming behaviors. Self-injured images are more widely spread among Tumblr users \citep{seko2018self} and exposure to them can lead to a risk for self-harm and suicide in vulnerable users \citep{doi:10.1177/1461444819850106}. While social media platforms are working on their explicit content moderation policies, a significant part of them remains undetected. At the same time, there are several positive narratives from self-injured survivors, which require a proactive stance to promote them \citep{seko2018self}. The majority of studies on automated detection of such content are based on textual, visual, and network content analysis: eating disorder \citep{10.1145/3018661.3018706}, self-harm \citep{losada2020overview,parapar2021overview}, self-harm detection on textual, visual and social content \citep{10.1145/3038912.3052555}. We have not found any literature for automatic detection and analysis of self-inflicted harmful memes.  

\begin{table*}[!t]
\centering
\setlength{\tabcolsep}{2pt}
\scalebox{0.65}{
\begin{tabular}{@{}llllclrrr@{}}
\toprule
\multicolumn{1}{c}{\textbf{Types}} & \multicolumn{1}{c}{\textbf{Publication}} & \multicolumn{1}{c}{\textbf{Task}} & \multicolumn{1}{c}{\textbf{Dataset}} & \textbf{Cl. T} & \multicolumn{1}{c}{\textbf{Approach}} & \multicolumn{1}{c}{\textbf{AUC}} & \multicolumn{1}{c}{\textbf{Acc.}} & \multicolumn{1}{c}{\textbf{F1}} \\ \midrule
\multirow{3}{*}{Harm} & \multirow{3}{*}{\citep{pramanick-etal-2021-momenta-multimodal}} & Y/N & \multirow{3}{*}{HarMeme} & B & \multirow{3}{*}{VisualBERT} &  & 0.81 & 0.80 \\
 &  & VH/Ph/NH &  & M &  &  & 0.74 & 0.54 \\
 &  & Tar. Ident. &  & M &  &  & 0.76 & 0.66 \\ \cmidrule(l){2-9}
\multirow{6}{*}{Harm} & \multirow{6}{*}{\citep{pramanick-etal-2021-momenta-multimodal}} & Y/N & \multirow{3}{*}{Harm-C} & B & \multirow{6}{*}{\begin{tabular}[c]{@{}l@{}}MOMENTA: \\ CLIP, VGG-19, \\ DistilBERT, \\ CMAF\end{tabular}} &  & 0.84 & 0.83 \\
 &  & VH/Ph/NH &  & M &  &  & 0.77 & 0.55 \\
 &  & Tar. Ident. &  & M &  &  & 0.78 & 0.70 \\ 
 &  & Y/N & \multirow{3}{*}{Harm-P} & B &  &  & 0.90 & 0.88 \\
 &  & VH/Ph/NH &  & M &  &  & 0.87 & 0.67 \\
 &  & Tar. Ident. &  & M &  &  & 0.79 & 0.69 \\ \midrule 
\multirow{2}{*}{Hate} & \multirow{2}{*}{\citep{zia-etal-2021-racist}} & PC & \multirow{2}{*}{FBHM} & ML & \multirow{2}{*}{\begin{tabular}[c]{@{}l@{}}CIMG, CTXT\\LASER, LaBSE\end{tabular}} & 0.96 &  &  \\
 &  & PC. AT. &  & ML &  & 0.97 &  &  \\ \cmidrule(l){2-9}
\multirow{4}{*}{Hate} & \multirow{4}{*}{\citep{Chandra_Jew2021}} & \multirow{2}{*}{Antisemitism} & Gab & \multirow{2}{*}{B} & \multirow{4}{*}{MFAS} &  & 0.91 &  \\ 
 &  &  & Twitter &  &  &  & 0.71 &  \\
 &  & \multirow{2}{*}{\begin{tabular}[c]{@{}l@{}}Antisemitism\\Category\end{tabular} } & Gab & \multirow{2}{*}{M} &  &  & 0.67 &  \\
 &  &  & Twitter &  &  &  & 0.68 &  \\ \cmidrule(l){2-9}
\multirow{2}{*}{Hate} & \multirow{2}{*}{\citep{kirk-etal-2021-memes}} & \multirow{2}{*}{Hateful} & FBHM & \multirow{2}{*}{B} & \multirow{2}{*}{CLIP} &  &  & 0.56 \\
 &  &  & Pinterest &  &  &  &  & 0.57 \\ \cmidrule(l){2-9}
\multirow{2}{*}{Hate} & \multirow{2}{*}{\citep{Lee2021Disen}} & \multirow{2}{*}{Hateful} & FBHM & \multirow{1}{*}{B} & \multirow{2}{*}{DisMultiHate} & 0.83 & 0.76 &  \\
 &  &  & MultiOFF &  &  &  &  & 0.65 \\  \cmidrule(l){2-9}
Hate & \citep{Gomez2020exploring} & Hatespech & MMHS150K & \multirow{1}{*}{B} & FCM, Inception-V3, LSTM & 0.73 & 0.68 & 0.70 \\  \cmidrule(l){2-9} 
Hate & \citep{fersini2019detecting} & Sexist & The MEME & \multirow{1}{*}{B} & Late fusion &  &  & 0.76 \\ \cmidrule(l){2-9}
Hate & \citep{sabat2019hate} & Hateful & Google & B & BERT, VGG-16, MLP &  & 0.83 & \multirow{2}{*}{} \\ \midrule

\multirow{2}{*}{Off.} & \multirow{2}{*}{\citep{AOMD2021}} & \multirow{2}{*}{Offensive} & Gab & \multirow{2}{*}{B} & \multirow{2}{*}{\begin{tabular}[c]{@{}l@{}}Faster R-CNN, ResNet50, \\ Glove-based LSTM, BERT, MLP\end{tabular}} &  & 0.69 & 0.56 \\ 
 &  &  & Reddit &  &  &  & 0.72 & 0.49 \\ \cmidrule(l){2-9}
\multirow{2}{*}{Off.} & \multirow{2}{*}{\citep{9582340}} & \multirow{2}{*}{Offensive} & Reddit &  \multirow{2}{*}{B} & \multirow{2}{*}{\begin{tabular}[c]{@{}l@{}}YOLO V4, ConceptNET, GNN\end{tabular}} &  & 0.73 & 0.49 \\
 &  &  & Gab & &  &  & 0.70 & 0.55 \\ \cmidrule(l){2-9}

\multirow{2}{*}{Off.} & \multirow{2}{*}{\citep{9402406}} & Offensive & \multirow{2}{*}{Off. Int.} & \multirow{1}{*}{B} & \begin{tabular}[c]{@{}l@{}}CNN, GloVe, LSTM\end{tabular} &  & 0.71 & \multirow{2}{*}{} \\
 &  & Off. Int. &  & \multirow{1}{*}{M} & \begin{tabular}[c]{@{}l@{}}CNN, FastText, LSTM\end{tabular} &  & 0.99 &  \\ \cmidrule(l){2-9}
 
Off. & \citep{suryawanshi-etal-2020-multimodal} & Offensive & MultiOFF & \multirow{1}{*}{B} & \begin{tabular}[c]{@{}l@{}}Early fusion: Stacked LSTM, \\ BiLSTM/CNN-Text,  VGG16\end{tabular} &  &  & 0.50 \\ \midrule

Prop. & \citep{dimitrov2021detecting} & \begin{tabular}[c]{@{}l@{}}Prop. Tech.\end{tabular} & \multirow{1}{*}{Facebook} & ML & VisualBERT &  &  & 0.48 \\ \cmidrule(l){2-9}

Prop. & \citep{tian-etal-2021-mind} & Prop. Tech.: (T) & \multirow{1}{*}{Facebook} & ML & \begin{tabular}[c]{@{}l@{}}Ensemble: BERT, RoBERTa, \\XLNet, ALBERT, DistilBERT, \\ DeBERTa, Char n-gram\end{tabular} &  &  & 0.59 \\ \cmidrule(l){2-9}
Prop. & \citep{gupta-etal-2021-volta} & Prop. Tech.: (S) & \multirow{1}{*}{Facebook} & ML & RoBERTa &  &  & 0.48 \\ \cmidrule(l){2-9}
Prop. & \citep{feng-etal-2021-alpha} & Prop. Tech. & \multirow{1}{*}{Facebook} & ML & RoBERTa, Embeddings &  &  & 0.58 \\ \midrule

CB & \citep{Hosseinmardi2015cyberbull} & CB Inci.  & Instagram & \multirow{2}{*}{B} & \begin{tabular}[c]{@{}l@{}}SVD +(Unigram, 3-gram), \\ kernelPCA+ meta data, lin. SVM\end{tabular} &  & 0.87 &  \\ \cmidrule(l){2-9}

\multirow{4}{*}{CB} & \multirow{4}{*}{\citep{suryawanshi-etal-2020-dataset}} & \multirow{4}{*}{Troll} & \multirow{4}{*}{TamilMemes} &  & ResNet (Tr: TM) &  &  & 0.52 \\
 &  &  &  & \multirow{2}{*}{B} & ResNet (Tr: TM + iNet) &  &  & 0.52 \\
 &  &  &  &  & MobileNet (Tr.: TM + iNet + Fl1k) &  &  & 0.47 \\
 &  &  &  & B & ResNet (Tr.: TM + iNet + Fl30k) &  &  & 0.52 \\ \bottomrule
\end{tabular}
}
\caption{Summary of the experimental results for the automatic detection of harmful memes. Y/N: positive and negative class labels; VH: Very harmful, PH: Partially-harmful, NH: Non-harmful; Tar. Ident.: Target Identification; PC: Protected category identification; PC. AT. : Protected category attack type; Off. Int.: Offense intensity prediction; Off: Offensive; Prop.: Propaganda; Prop. Tech.: Propaganda techniques, Prop. Tech.: (T): Text, Prop. Tech.: (S): text span; CB Inci.: Cyberbullying Incidents; CMAF: Cross-modal attention fusion. Cl.T: Classification task; B: Binary, M: Multi-class, ML: Multi-class and Multilabel; TM: TamilMemes, iNet: ImageNet, Fl: Flickr. 
{\bf More detail can be found at:} \href{http://github.com/firojalam/harmful-memes-detection-resources}{http://github.com/firojalam/harmful-memes-detection-resources}.
}
\label{tab:related_studies_exp_results}
\end{table*}

\subsection{Summary}

Table \ref{tab:related_studies_exp_results} summarizes the state of the art for automatic detection of different types of harmful memes, exploring different tasks, datasets, and approaches. In the majority of these studies, the tasks are formulated in a binary setting. While the outcome of a binary setting is useful, multi-class and multi-label setttigs would be more desirable, e.g.,~as addressed in \citep{dimitrov2021detecting} for propaganda detection and protected category detection \citep{zia-etal-2021-racist}. The majority of the studies used state-of-the-art pre-trained visual (e.g., VGG and ResNet), NLP (e.g., BERT), and multimodal (e.g., Visual BERT and CLIP) models. Data augmentation and ensembles were used in several studies. Table~\ref{tab:related_studies_exp_results} shows variations of F1 such as micro, macro, and weighted; more details can be found at \href{https://github.com/firojalam/harmful-memes-detection-resources}{https://github.com/firojalam/harmful-memes-detection-resources}. Overall, the results are comparatively better for harmful and for hateful memes than for the remaining tasks. For binary classification tasks like troll identification, the results are only slightly better than random, which highlights the complexity of these tasks.

\section{Repackaging Memes for Harmful Agendas}

Repackaging via remixing or mimicking the meme is a common practice facilitating their adoption across languages and cultures \citep{shifman2013memes}, which often imply harm. For example, popular memes are often repackaged with misogynistic intent. Common ideas that mock specific female identities include \emph{the terrible wife} or 
\emph{the crazy girlfriend}. For example, the \emph{Distracted Boyfriend} meme\footnote{\url{https://knowyourmeme.com/memes/distracted-boyfriend}} has been repackaged many times with varying intent, including harm and humor.

Another example is the \emph{Proud Boys} meme\footnote{\url{https://www.populismstudies.org/wp-content/uploads/2021/03/ECPS-Organisation-Profile-Series-1.pdf}}, which has peculiar characteristics. Its proponents work in gangs, indulge in violence and alcohol, follow a uniform code for appearances and collectively accepted logos to depict their identity. 

The use of \emph{Pepe the Frog} reinstates their deeply rooted affiliation to far-right ideologies. Their version of Pepe is a variation that depict him donning the Proud Boys uniform (black Fred Perry polo with gold trim), whilst displaying the OK hand gesture.

\section{Cultural Influence and Multilinguality}

\citet{shifman2013memes} introduced the term \emph{user-generated globalization}, which refers to translation, customization, and distribution of memes across the globe by ordinary online users. In particular, they studied a joke related to computers and romantic relations and its traslated version in the top nine non-English languages and found that the joke adapted very well in most of these languages, except for Arabic, which might be due to culture-specific inappropriateness. They further found limited localization of the joke in Chinese, German, and Portuguese.

A recent study \citep{Jordan_multinationalmeme} found that most memes either pre-dominantly belonged to anglophone organizations or were derived from anglophone references like the ``One Does Not Simply Walk Into Mordor'' meme, which appeared in Germany's Ein Prozent. Most European organizations leverage different genres of images like \emph{share-posts} and \emph{templates} specifically designed for online circulation and orthogonal to the irreverent and participatory nature of memes. In addition to the localized cultural adaptation and customization, memes can use multiple languages. Such examples can be found in the TamilMemes dataset \citep{suryawanshi-etal-2020-dataset}. Modelling such memes is complex, as is evident from the results reported in \citep{hegde2021images}.

\section{Major Challenges}

\noindent \textbf{$\bullet$ Complex abstraction:} One key advantage of memes is their efficacy to abstract away complex ideas using creative and powerful customization of visual and linguistic nuances. At the same time, memes with overlapping snippets, patterned text and irony, sarcasm or implicit anti-semitism are non-trivial \citep{Chandra_Jew2021}. For instance, the subtle usage of triple parentheses in memes can insinuate a targeted entity whilst underlining an anti-semitic narrative \citep{tuters_para2020}. Moreover, \emph{sexist memes} can promote casual sexism, disguised as humor, irony, sarcasm, and mockery \citep{Siddiqi2018}. This multi-layering of influential notions via multimodality poses major challenges for automatic meme analysis and requires sophisticated multimodal fusion to understand novel digital vernaculars.

\noindent \textbf{$\bullet$ Subjectivity in the annotation:} Subjective perceptions play a significant role for memes as a consequence of the complex interplay between the visual and the linguistic content, complemented by the lack of context~\citep{sep-perception-problem}. Moreover, harmful memes, which are prominently used for propaganda warfare, violate one's logic and rational thought. This reverberates as conflicting opinions during data collection and annotation. As noted in \citep{suryawanshi-etal-2020-multimodal}, uninitiated annotators were observed to incorrectly mark memes as offensive simply if their sentiments were hurt. This was also concluded from a user study in \citep{france_mys2021}, wherein out of 59 ambiguous misogynistic memes, only 23\% were correctly identified by crowd-sourced workers, while domain experts achieved 77\% expert agreement.

\noindent \textbf{$\bullet$ Inadequate solutions:}
Understanding the visual content in memes requires sophisticated solutions, as conventional approaches rely too much on hand-crafted features like low-level grey-scaling, coloured, photographic, and semantic features, along with ineffective modeling \citep{fersini2019detecting}. This is amplified by the predominantly non-discriminatory nature of visual descriptors in memes, emphasising textual and discourse-intensive modeling \citep{9582340,AOMD2021}. Visual clustering techniques such as pHash used for memes depicting standardized imagery like popular alt-right figures (e.g., \emph{Lauren Southern}, \emph{Richard Spencer}), as well as alt-right memes such as \emph{Pepe the Frog}, and anti-semitic ones such as the \emph{Happy Merchant} are insufficient to model the visual role-play, indicating the need for sophisticated visual analysis \citep{Jordan_multinationalmeme,zannetou2018}. 

\noindent \textbf{$\bullet$ Insufficient sample size:}  
Meme analysis requires a rich set of features and meta-data, which in turn needs a sample size large enough to be generalizable at scale \citep{NatourAboriginals2020}. Similarly, a keyword-based platform-dependent collection of memes could yield a biased representation of the sample space, and hence could over-represent typical memetic characteristics \citep{Fairchild2020ItsFB}.   

\noindent \textbf{$\bullet$ Rapid evolution}: Harmful memes evolve quickly, fueled by new events or by malicious adversaries looking for new ways to bypass existing online detection systems. While humans can generally use prior knowledge to understand new harmful concepts and tasks by looking at a few examples, AI systems struggle to generalize well from a few examples \citep{wang2020generalizing}. Few-shot learning (FSL) is a new machine learning paradigm that has recently shown breakthrough results in NLP \citep{brown2020language} and vision tasks \cite{fan2021generalized}. It is crucial to advance FSL in the multimodal domain to adapt rapidly and to recognize new evolving types of harmful memes \citep{tsimpoukelli2021multimodal,tejankar2021fistful}. Unlike traditional AI that mainly relies on pattern-matching with labeled data, FSL-based AI systems can evolve to new harmful memes and policies using a handful of examples and can take action immediately instead of waiting for months for the labeled data to be collected.

\noindent \textbf{$\bullet$ Contextualization:} Understanding many memes requires complex and multimodal reasoning that is based upon a certain contextual background, which may span over diverse levels of abstraction, such as \emph{common sense} \citep{9582340}, \emph{factual} \citep{zhu2020enhance}, and \emph{situational} \citep{sabat2019hate}. This contextual information may be conveyed both independently and jointly via textual and visual cues. Analyzing this information can be crucial, but it is often not explicitly available for the target meme. 

\noindent \textbf{$\bullet$ Platform restrictions:} The non-standardization of \textit{user accountability and transparency} across constantly evolving social networking services has posed challenges for the systematic study of online harm-detection. For example, the freedom of being anonymous has obscured racial integrity and accountability, effectively complicating harmful discourse analysis \citep{Nikolas2016}. Moreover, the complex designs and governance policies of platforms such as WhatsApp meant that they focused on their \emph{secure} but unabated use for disseminating systematic racism \citep{Fernandez_2020}. As observed by \citet{zannetou2018}, the investigation of an actively evolving community like Gab, using a Hawkes process, might err the observations \citep{zannetou2018}.

\noindent \textbf{$\bullet$ Identifying real instigators of harm:}
Poe's law emphasizes the understanding of the actual intent while distinguishing between online satire and extremism \citep{studamerhumor2019}. Similar ambiguity could also be observed while distinguishing between the real faces of white supremacy and its participatory audience \citep{studamerhumor2019}. Interestingly, memes like \emph{triple parenthesis} can render the targets obscure \citep{tuters_para2020}. Even the regulatory bodies find it challenging to clearly distinguish between anti-democracy extremists and anti-democratic alt-right factions \citep{nordinNeoNaziAltright2021}. Consequently, one must also be careful while associating the alt-right with culture. It is instead a historical phenomenon that leverages culture as a tool for its propagation \citep{Dafaure2020TheM}.

\section{Future Forecasting}

\noindent \textbf{$\bullet$ Characterizing vehicles of harm:} 
Satire is not only used as a progressive tool to resist bigotry, but it is also weaponized by malicious actors totowards high-jacking the online discussions \citep{studamerhumor2019}. It is thus important to decode the discourse and understand the communication that memes are part of \citep{proudboys2018}. Exploring the points of the confluence of youth with far-right memes will help highlight where and how messages of extreme violence circulate and transit back and forth between malicious actors and receptive users \citep{askaniusMurder2021}. It could also be insightful to examine how symptomatic the discourse rhetoric of the anecdotal reference is, within the backdrop of rooted antisemitic perspectives, like the nebulous \emph{Othering}.

\noindent \textbf{$\bullet$ Cross-cultural studies:} Systems would require to factor in the prejudices and the stereotypes surrounding various minorities for being sensitive towards racially hateful memes. One hypothesis is that the relationship between offline micro-aggression and online perception of racism will become more prominent in settings where Whites are not the majority. This presents the scope of investigating cross-cultural and cross-contextual implications for the racism experienced and perceived online \citep{WILLIAMS2016424}.

\noindent \textbf{$\bullet$ Empirical \textit{in addition to} theoretical:} There are few compelling questions arising from the existing understanding of harmful memes regarding the cause of their potency to instigate harm, cross-platform transitioning and outcomes. Few of them being: To what extent are the ``hate jokes'' part of the slow yet steady process of normalizing online extremism in mainstream media? What are the consequences of transitioning from their original lair to the mainstream? What is the reaction of the general public when exposed to such content~\citep{nordinNeoNaziAltright2021}? Clearly, the assessment of the prevalence of different visual forms like memes, photography, and artwork in online communications, along with the cryptic use of visual-linguistic semiotics, requires active empirical investigations \citep{Jordan_multinationalmeme}.

\noindent \textbf{$\bullet$ Rich metadata:} The use of enriching features such as the tags associated with the social media posts, incorporating video data along with contextual information like user profiles \citep{Chandra_Jew2021}, and using intermediate representations to capture higher levels of abstractions that leverage both the image and the text modalities can help model complex tasks. Moreover, the contextual knowledge supplementing such abstract information becomes indispensable for automated meme analysis \cite{9582340}.

\noindent\textbf{$\bullet$ Multi-class and multi-label classification:} 
As highlighted in Table~\ref{tab:related_studies_exp_results}, the existing classification setups are primarily binary. However, a more fine-grained multi-class and multi-label setup can enhance the decision-making process, as required in many scenarios. For example, a meme labeled as hateful~\citep{kiela2020hateful}, which has the characteristics of violence and misogyny, loses its specificity. Attempts in this direction include fine-grained analysis of hateful~\citep{zia-etal-2021-racist} and propagandistic \citep{dimitrov2021detecting} memes, detecting the victims targeted by harmful memes \citep{pramanick-acl,DISARM:2022}, and understanding who is the hero, the villain, and the victim \citep{sharma-etal-2022-findings}.

\noindent \textbf{$\bullet$ Memetic moderation:} Counter-narratives can help address the selective targeting via harmful memes \citep{blmBecky2020}. The utility of the post-modern transgression and humor must not be left to the alt-right extremists just because they were successful in weaponizing them, as essentially it reinstates their belief that the ``left can't meme'' \citep{Dafaure2020TheM}. Creating counter memes can help raise awareness about racial issues \citep{Yoon2016WhyII}. Reclaiming the digital space and indulging in subversive reactions by leveraging the participatory humor using \emph{digilanties} (online vigilantes) can help mitigate the collective menace impended by the systematic and subtle oppression of women \citep{Drakett2018}.

\section{Conclusion}
\label{sec:conclusion}

We presented a survey of the current intelligent technologies for detecting and understandingh harmful memes. Based on a systematic analysis of recent literature, we first proposed a new typology of harmful memes, and then we highlighted and summarized the relevant state of the art. We then discussed the lessons learned and the major challenges that need to be overcome. Finally, we suggested several research directions, which we forecast will emerge in the near future.

\section*{Acknowledgments}
The work was partially supported by a Wipro research grant, Ramanujan Fellowship, the Infosys Centre for AI, IIIT Delhi, and ihub-Anubhuti-iiitd Foundation, set up under the NM-ICPS scheme of the Department of Science and Technology, India. It is also part of the Tanbih mega-project, which is developed at the Qatar Computing Research Institute, HBKU, and aims to limit the impact of ``fake news,'' propaganda, and media bias by making users aware of what they are reading.

{\small
\bibliographystyle{named}
\bibliography{all_bib}
}
\end{document}